\documentclass{article}

\usepackage{microtype}
\usepackage{graphicx}
\usepackage{subfigure}
\usepackage{booktabs} 

\usepackage{hyperref}

\usepackage[accepted]{icml2021}

\usepackage[noend]{algpseudocode}
\usepackage{subfigure, multirow}
\usepackage{booktabs}
\usepackage{array}
\usepackage{amsmath}
\usepackage{diagbox}
\usepackage{xspace}
\usepackage{cleveref}
\usepackage{bm}
\usepackage[ruled,linesnumbered,vlined]{algorithm2e}
\usepackage{graphicx}
\usepackage{floatrow}
\newfloatcommand{capbtabbox}{table}[][\FBwidth]
\usepackage{pgfplots}
\usepackage{wrapfig}
\usepackage{ragged2e}
\usepackage{xcolor}

\newcommand{\bx}{\bm{x}}

\newcommand{\bs}{\bm{s}}
\newcommand{\btheta}{\bm{\theta}}

\newcommand{\mt}{\mathcal{T}}
\newcommand{\ms}{\mathcal{S}}
\newcommand{\ml}{\mathcal{L}}
\newcommand{\ma}{\mathcal{A}}

\makeatletter
\DeclareRobustCommand\onedot{\futurelet\@let@token\@onedot}
\def\@onedot{\ifx\@let@token.\else.\null\fi\xspace}
\def\eg{\emph{e.g}\onedot} 
\def\ie{\emph{i.e}\onedot}

\renewcommand{\paragraph}{%
	\@startsection{paragraph}{4}{\z@}%
	{0.1em \@plus 0.5ex \@minus 0.2ex}{-1em}%
	{\normalsize\bf}%
}
\makeatother

\usepackage{amsmath,amsfonts,bm}

\def\eqref#1{equation~\ref{#1}}

\def\1{\bm{1}}

\DeclareMathAlphabet{\mathsfit}{\encodingdefault}{\sfdefault}{m}{sl}
\SetMathAlphabet{\mathsfit}{bold}{\encodingdefault}{\sfdefault}{bx}{n}

\begin{document}

\twocolumn[
\icmltitle{Dataset Condensation with Differentiable Siamese Augmentation}




\begin{icmlauthorlist}
\icmlauthor{Bo Zhao}{to}
\icmlauthor{Hakan Bilen}{to}
\end{icmlauthorlist}

\icmlaffiliation{to}{School of Informatics, The University of Edinburgh, UK}

\icmlcorrespondingauthor{Bo Zhao}{bo.zhao@ed.ac.uk}
\icmlcorrespondingauthor{Hakan Bilen}{hbilen@ed.ac.uk}


\vskip 0.3in
]



\printAffiliationsAndNotice{}  

\begin{abstract}
In many machine learning problems, large-scale datasets have become the de-facto standard to train state-of-the-art deep networks at the price of heavy computation load. In this paper, we focus on condensing large training sets into significantly smaller synthetic sets which can be used to train deep neural networks from scratch with minimum drop in performance. Inspired from the recent training set synthesis methods, we propose Differentiable Siamese Augmentation that enables effective use of data augmentation to synthesize more informative synthetic images and thus achieves better performance when training networks with augmentations. Experiments on multiple image classification benchmarks demonstrate that the proposed method obtains substantial gains over the state-of-the-art, 7\% improvements on CIFAR10 and CIFAR100 datasets. We show with only less than 1\% data that our method achieves 99.6\%, 94.9\%, 88.5\%, 71.5\% relative performance on MNIST, FashionMNIST, SVHN, CIFAR10 respectively. We also explore the use of our method in continual learning and neural architecture search, and show promising results.
\end{abstract}

\section{Introduction}
\label{sec:intro}
Deep neural networks have become the go-to technique in several fields including computer vision, natural language processing and speech recognition thanks to the recent developments in deep learning~\cite{krizhevsky2012imagenet, simonyan2014very, szegedy2015going, he2016deep} and presence of large-scale datasets~\cite{deng2009imagenet, lin2014microsoft, antol2015vqa, abu2016youtube}.
However, their success comes at a price, increasing computational expense, as the state-of-the-art models have been primarily fueled by larger models (\eg \cite{devlin2018bert,radford2019language, dosovitskiy2021image} and massive datasets~(\eg \cite{OpenImages,Chen20,kwiatkowski2019natural}). 
For example, it takes 12.3k TPU days to train EfficientNet-L2~\cite{xie2020self}  on JFT-300M dataset~\cite{sun2017revisiting}. 
To put in a perspective, the energy consumption for training EfficientNet-L2 once is about $3\times 10^7$ J, assuming that the TPU training power is 100W. 
Ideally, the same energy is sufficient to launch a $30$ kg object to the outer space, \ie reaching Kármán line which costs gravitational potential energy $10^6$ J/kg.
More dramatically, the computational cost significantly increases when better neural architectures are searched and designed due to many trials of training and validation on the dataset for different hyper-parameters \cite{bergstra2012random,elsken2019neural}.
Significantly decreasing these costs without degrading the performance of the trained models is one of the long-standing goals in machine learning \cite{agarwal2004approximating}.
To address these challenges, this paper focuses on reducing the training data size by learning significantly smaller synthetic data  to train deep neural networks with minimum drop in their performance.

The standard way to reduce the training set size is to use a smaller but equally informative portion of data, namely a \emph{coreset}. 
In literature, there is a large body of coreset selection methods for various target tasks, \eg accelerating model training in neural architecture search \cite{shleifer2019using, such2020generative}, storing previous knowledge compactly in continual learning \cite{rebuffi2017icarl, toneva2019empirical} and efficient selection of samples to label in active learning \cite{sener2017active}.
However, their selection procedures rely on heuristics and thus do not guarantee any optimal solution for the downstream tasks (\eg image classification).
In addition, finding such an informative coreset may not always be possible when the information in the dataset is not concentrated in few samples but uniformly distributed over all of them. 

Motivated by these shortcomings, a recent research direction, \emph{training set synthesis} aims at \emph{generating} a small training set which is further used to train deep neural networks for the downstream task~\citep{wang2018dataset, sucholutsky2019soft, bohdal2020flexible, such2020generative, nguyen2021dataset, zhao2021dataset}. 
In particular, Dataset Distillation~(DD)~\cite{wang2018dataset} models the network parameters as a function of synthetic training data, and then minimize the training loss on the real training data by optimizing the synthetic data.
\citet{sucholutsky2019soft} extend DD by learning synthetic images and soft labels simultaneously.
\citet{bohdal2020flexible} simplify DD by only learning the informative soft labels for randomly selected real images.
\citet{such2020generative} propose to use a generator network instead of directly learning synthetic data. 
\citet{nguyen2021dataset} reformulates DD in a kernel-ridge regression which has a closed-form solution. 
\citet{zhao2021dataset} propose Dataset Condensation~(DC) that ``condenses'' the large training set into a small synthetic set by matching the gradients of the network parameters w.r.t. large-real and small-synthetic training data. 
The authors show that DC can be trained more efficiently by bypassing the bi-level optimization in DD while significantly outperforming DD in multiple benchmarks.
Despite the recent success of the training set synthesis over the coreset techniques, especially in low-data regime, there is still a large performance gap between models trained on the small synthetic set and those trained on the whole training set.
For instance, models that are trained on DD and DC synthetic sets obtain 38.3\% and 44.9\% accuracy respectively with 10 images per class on the CIFAR10 dataset, while a model trained on the whole dataset (5000 images per class) obtains 84.8\% .


An orthogonal direction to increase data efficiency and thus generalization performance is data augmentation, a technique to expand training set with semantic-preserving transformations
\cite{krizhevsky2012imagenet, zhang2018mixup, yun2019cutmix, chen2020simple, chen2020exploring}. 
While they can simply be used to augment the synthetic set that are obtained by a training set synthesis method, we show that naive strategies lead to either drops or negligible gains in performance in \Cref{sec:exp}. 
This is because the synthetic images i) have substantially different characteristics from natural images, ii) are not learned to train deep neural network under various transformations.
Thus we argue that an effective combination of these two techniques is non-trivial and demands careful data augmentation design and principled learning procedure.

\begin{figure*}[t]
    \centering
    \includegraphics[width=0.8\linewidth]{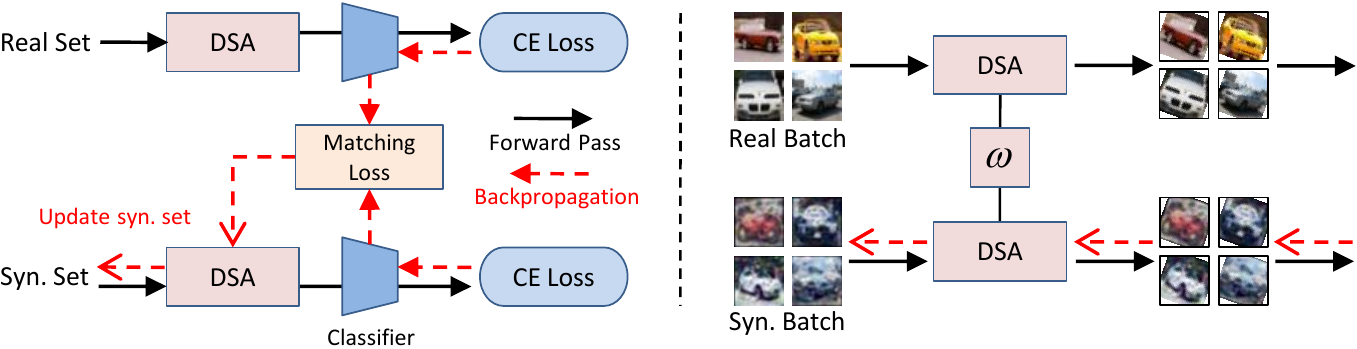}
    \caption{\footnotesize{Dataset condensation with differentiable Siamese augmentation. Differentiable Siamese augmentation (DSA) applies the same parametric augmentation (\eg rotation) to all data points in the sampled real and synthetic batches in a training iteration. The gradients of network parameters w.r.t. the sampled real and synthetic batches are matched for updating the synthetic images. {A DSA example is given that rotation with the same degree is applied to the sampled real and synthetic batches.}
    }}
    \label{fig:method}
    \vspace{-12pt}
\end{figure*}

In this paper we propose a principled method to enable learning a synthetic training set that can be effectively used with data augmentation to train deep neural networks.
Our main technical contribution is \emph{Differentiable Siamese Augmentation}~(DSA), illustrated in \Cref{fig:method}, that applies the same randomly sampled data transformation to both sampled real and synthetic data at each training iteration and also allows for backpropagating the gradient of the loss function w.r.t. the synthetic data by differentiable data transformations. 
{Applying various data transformations (\eg $15^{\circ}$ clockwise rotation) simultaneously to both real and synthetic images in training has three key advantages.
First our method can exploit the information in real training images more effectively by augmenting them in several ways and transfer this augmented knowledge to the synthetic images.
Second sharing the same transformation across real and synthetic images allows the synthetic images to learn certain prior knowledge in the real images (\eg the objects are usually horizontally on the ground).}
Third, most importantly, once the synthetic images are learned, they can be used with various data augmentation strategies to train different deep neural network architectures.
We validate these advantages in multiple image classification benchmarks and show that our method significantly outperforms the state-of-the-art with a wide margin, around 7\% on CIFAR10/100 datasets\footnote{The implementation is available at \url{https://github.com/VICO-UoE/DatasetCondensation}.}.
Finally we explore the use of our method in continual learning and neural architecture search, and show promising results.


\section{Related Work}
\label{sec:relatedwork}


In addition to the coreset selection and training set synthesis methods that are discussed in \cref{sec:intro}, our method is also related to data augmentation techniques and Generative Adversarial Networks (GANs).

\paragraph{Data Augmentation.} Many deep neural networks adopts data transformations for expanding the effective training set size, reducing overfitting and thus improving their performance. 
The most popular augmentation strategies include color jittering \cite{krizhevsky2012imagenet}, cropping~\cite{krizhevsky2012imagenet}, cutout~\cite{devries2017improved}, flipping, scale, rotation.
More elaborate augmentation strategies are Mixup \cite{zhang2018mixup} and CutMix \cite{yun2019cutmix}.
These augmentation strategies are typically applied to various image recognition problems where the label is invariant to transformations of the input and the transformations do not have to be be differentiable w.r.t. the original input image.
While we also use several data augmentation techniques, our focus is to synthesize training images that results in gradients that are equivariant to the ones from real images.
In addition, we use differentiable augmentations such that gradients can go through augmentation function and back-propagate to synthetic data.


\paragraph{Auto-augmentation.} This line of work investigates how to automatically find the best augmentation strategy instead of manually designing by either learning a sequence of transformation functions in an adversarial optimization~\cite{ratner2017learning}, using a reinforcement learning algorithm~\citet{cubuk2019autoaugment}, or learning the parameters of parametric feature augmentations~\cite{yan2020augmented}. 
In contrast, our goal is not to find the best augmentation for training data but to synthesize the training data that is equipped with the augmentation ability for the downstream task.

\paragraph{GANs \& Differentiable Augmentation.} GANs~\cite{goodfellow2014generative, mirza2014conditional, radford2015unsupervised} typically aim at generating real-looking novel images by fooling a discriminator network.
Differentiable Augmentation~\cite{zhao2020differentiable, tran2020data, zhao2020image, karras2020training} has recently been applied for improving their training and in particular for preventing discriminators from memorizing the limited training set.
Though they also apply differentiable augmentation to both real and fake images, augmentations are independently applied to real and fake ones. 
In contrast we use a Siamese augmentation strategy which is explicitly coordinated to apply the same transformation to both real and synthetic images.
In addition, our goal, which is to generate a set of training data that can be used to efficiently train deep neural networks from scratch, differs from theirs and our images do not have to be realistic.

\section{Method}
\label{sec:method}
Here we first review DC~\cite{zhao2021dataset}, then describe the proposed our DSA method and its training algorithm. 


\subsection{Dataset Condensation Review}
\label{sec:dc}
Assume that we are given a large training set $\mt=\{(\bx_1,y_1),\dots,(\bx_{|\mt|},y_{|\mt|})\}$ with $|\mt|$ image and label pairs.
DC~\cite{zhao2021dataset} aims at learning a much smaller set with $|\ms|$ synthetic image and label pairs $\ms=\{(\bs_1,y_1),\dots,(\bs_{|\ms|},y_{|\ms|})\}$ from $\mt$ such that a deep network trained on $\ms$ obtains comparable generalization performance to a deep neural network that is trained on $\mt$.
Let $\phi_{\btheta^\mt}$ and $\phi_{\btheta^\ms}$ denote the deep neural networks with parameters $\btheta^{\mt}$ and $\btheta^{\ms}$ that are trained on $\mt$ and $\ms$ respectively.
The goal of DC can be formulated as:
\begin{equation}
    \mathbb{E}_{\bx \sim P_{\mathcal{D}}}[\ell(\phi_{\btheta^\mt}(\bx),y)]\simeq\mathbb{E}_{\bx \sim P_{\mathcal{D}}}[\ell(\phi_{\btheta^\ms}(\bx),y)] 
    \label{eq:goalDC}
\end{equation}
over the real data distribution $P_{\mathcal{D}}$ with loss function $\ell$ (\ie cross-entropy loss) .
In practice, their generalization performances are measured on an unseen test set.


A possible way to achieve the comparable performance in \cref{eq:goalDC} is obtaining a similar solution to $\btheta^\mt$, after the parameters of the network are trained on $\ms$, \ie $\btheta^\ms\approx\btheta^\mt$.
However, solving this w.r.t. $\ms$ involves nested loop optimization over network parameters $\btheta$ and synthetic data $\ms$ which is typically not scalable to large models and multi-step optimization.
Instead the authors in \cite{zhao2021dataset} hypothesize that a similar solution can be achieved, when the parameter updates for $\btheta^{\mt}_t$ and $\btheta^{\ms}_t$ are approximately equal at each training iteration $t$, given the same initialization $\btheta^{\mt}_0=\btheta^{\ms}_0$.
In addition, assuming that $\btheta^{\ms}_t=\btheta^{\mt}_t$ can be satisfied at each iteration, the authors simplify the learning by using a single neural network parameterized by $\btheta$ and propose the following minimization problem:
\begin{equation}
     \min_{\ms} D(\nabla_{\btheta}\ml(\ms,\btheta_{t}),\nabla_{\btheta}\ml(\mt,\btheta_{t})), 
\label{eq.optgrad}
\end{equation} where 
\[
\ml(\ms,\btheta_t)=\frac{1}{|\ms|}\sum\limits_{(\bs,y)\in\ms}\ell(\phi_{\btheta_t}(\bs),y)),
\]
\vspace{-0.3cm}
\[
\ml(\mt,\btheta_t)=\frac{1}{|\mt|}\sum\limits_{(\bx,y)\in\mt}\ell(\phi_{\btheta_t}(\bx),y))
\] and $D$ is a sum of cosine distances between the two gradients of weights associated with each output node at each layer. 
We refer the readers to \cite{zhao2021dataset} for more detailed explanation.




\subsection{Differentiable Siamese Augmentation (DSA)}
Here we explain how data augmentation strategies can be effectively used with DC formulation.
One naive way is to apply data augmentation to the synthetic images post-hoc, after they are learned. 
However, this strategy results in negligible performance gains (demonstrated in \Cref{sec:exp}), as the synthetic images are {not} optimized to be augmented. 
Hence, a more principled way is to apply data augmentation while learning the synthetic images, which can be formulated by rewriting \cref{eq.optgrad}: 
\begin{equation}
    \min_{\ms} D(\nabla_{\btheta}\ml(\ma(\ms,\omega^\ms), \btheta_{t}),\nabla_{\btheta}\ml(\ma(\mt,\omega^\mt), \btheta_{t})),
    \label{eq.optgradAug}
\end{equation} where $\ma$ is a family of image transformations that preserves the semantics of the input (\ie class label) such as cropping, color jittering, flipping that are parameterized with $\omega^\ms$ and $\omega^\mt$ for the synthetic and real training sets respectively.

\paragraph{Siamese Augmentation.} 
In the standard data augmentation $\omega$ is randomly sampled from a predetermined distribution $\Omega$ for each image independently.
However, randomly sampling both $\omega^\ms$ and $\omega^\mt$ is not meaningful in our case, as this results in ambiguous gradient matching problem in \cref{eq.optgrad}.
For instance, in case of cropping, this would require a particular region of a synthetic image to produce gradients matching to the ones that are generated from different crops of real image at different training iterations.
Hence this method results in an averaging affect on the synthetic images and loss of information.
To address this issue, we instead use the same transformations across the synthetic and real training sets, \ie $\omega^\ms=\omega^\mt$. 
Thus we use one symbol $\omega$ in the remainder of the paper.
As two sets have different number of images $|\ms|\ll |\mt|$ and there is no one-to-one correspondence between them, we randomly sample a single transformation $\omega$ and apply it to all images in a minibatch pair at each training iteration. This also avoids the averaging effect in a minibatch.
This strategy enables correspondence between the two sets (\eg between $15^{\circ}$ clockwise rotation of synthetic and real set) and a more effective way of exploiting the information in the real training images and distilling it to the synthetic images in a more organized way without averaging effect. 
We illustrate the main idea in \Cref{fig:method}.

\paragraph{Differentiable Augmentation.} 
Solving \cref{eq.optgradAug} for $\ms$ involves computing the gradient for the matching loss $D$ w.r.t. the synthetic images ${\partial D(\cdot)}/{\partial \ms}$ by backpropagation:
\[
\frac{\partial D(\cdot)}{\partial \ms} = \frac{\partial D(\cdot)}{\partial \nabla_{\btheta}\ml(\cdot)} \frac{\partial \nabla_{\btheta}\ml(\cdot)}{\partial \ma(\cdot)} \frac{\partial \ma(\cdot)}{\partial \ms}.
\]
Thus the transformation $\ma$ has to be differentiable w.r.t. the synthetic images $\ms$.
Traditionally transformations used for data augmentation are not implemented in a differentiable way, as optimizing input images is not their focus.
Note that all the standard data augmentation methods for images are differentiable and can be implemented as differentiable layers.
Thus, we implement them as differentiable functions for deep neural network training and allow the error signal to be backpropagated to the synthetic images.

\begin{algorithm*}
\KwIn{Training set $\mt$}
\justifying{
\textbf{Required}: Randomly initialized set of {synthetic samples $\ms$} for $C$ classes, 
probability distribution over randomly initialized weights $P_{\btheta_0}$, deep neural network $\phi_{\btheta}$, number of training iterations $K$, number of inner-loop steps $T$, number of steps for updating weights $\varsigma_{\btheta}$ and synthetic samples $\varsigma_{\ms}$
in each inner-loop step respectively, learning rates for updating weights $\eta_{\btheta}$ and {synthetic samples $\eta_{\ms}$},
differentiable augmentation $\ma_{\omega}$ parameterized with $\omega$, augmentation parameter distribution $\Omega$, random augmentation $\ma$.}

\For{$k = 0, \cdots, K-1$}
{
	Initialize $\btheta_0 \sim P_{\btheta_0}$
	
	\For{$t = 0, \cdots, T-1$}
	{
	    
	    \For{$c = 0, \cdots, C-1$}  
		{
    		Sample $\omega\sim \Omega$ and a minibatch pair $B^\mt_c\sim\mt$ and $B^\ms_c\sim\ms$
    		\algorithmiccomment{$B^\mt_c$, $B^\ms_c$ are of class $c$.}
    
    		Compute $\ml^\mt_c=\frac{1}{|B^\mt_c|}\sum_{(\bx,y)\in B^\mt_c}\ell(\phi_{\btheta_t}(\ma_{\omega}(\bx)),y)$ and $\ml^\ms_c=\frac{1}{|B^\ms_c|}\sum_{(\bs,y)\in B^\ms_c}\ell(\phi_{\btheta_t}(\ma_{\omega}(\bs),y)$
    		
    		{Update $\ms_c\leftarrow \texttt{sgd}_{\ms}(D(\nabla_{\btheta}\ml^\ms_c(\btheta_{t}),\nabla_{\btheta}\ml^\mt_c(\btheta_{t})),\varsigma_{\ms},\eta_{\ms})$}
    		
		}
		Update $\btheta_{t+1} \leftarrow \texttt{sgd}_{\btheta}(\ml(\btheta_{t}, \ma(\ms)),\varsigma_{\btheta},\eta_{\btheta})$ \algorithmiccomment{Use $\ma$ for the whole $\ms$}
	}
}
\KwOut{$\ms$}
\caption{Dataset condensation with differentiable Siamese augmentation.}
\label{algo}
\end{algorithm*}


\subsection{Training Algorithm}
We adopt training algorithm in \cite{zhao2021dataset} for the proposed DSA, which is depicted in Alg.~\ref{algo}. 
To ensure that the generated synthetic images can train deep neural networks from scratch with any randomly initialized parameters, we use an outer loop with $K$ iterations where at each outer iteration we randomly initialize network parameters (\ie $\btheta_0\sim P_{\btheta_0}$) from a distribution $P_{\btheta_0}$ and train them from scratch.
In the inner loop $t$, we randomly sample an image transformation $\omega$ and a minibatch pair $B^\mt_c$, $B^\ms_c$ from the real and synthetic sets that contain samples from only class $c$, compute their average cross-entropy loss and gradients w.r.t. the model parameters separately.
Then we compute the gradient matching loss as in \cref{eq.optgradAug} and update the synthetic data $\ms_c$ by using stochastic gradient descent optimization with $\varsigma_{\ms}$ gradient descent steps and $\eta_{\ms}$ learning rate. 
{We repeat above steps for every class $c$ in the inner loop $t$.}
Alternatively, we update the model parameters $\btheta_t$ to minimize the cross-entropy loss on the augmented synthetic data with $\varsigma_{\btheta}$ gradient descent steps and $\eta_{\btheta}$ learning rate. 

\paragraph{Discussion.} We observe that using minibatches from multiple classes leads to a slower convergence rate in training.
The reason is that when the gradients $\nabla_{\btheta}\ml$ are averaged over samples from multiple classes, image/class correspondence for synthetic data is harder to retrieve from the gradients.


\section{Experiments}
\label{sec:exp}

\subsection{Datasets \& Implementation Details}
\paragraph{Datasets.} We evaluate our method on 5 image classification datasets, MNIST \cite{lecun1990optimal}, FashionMNIST \cite{xiao2017fashion}, SVHN \cite{netzer2011reading}, CIFAR10 and CIFAR100 \cite{krizhevsky2009learning}. 
Both the MNIST and FashionMNIST datasets have 60,000 training and 10,000 testing images of 10 classes. 
SVHN is a real-world digit dataset which has 73,257 training and 26,032 testing images of 10 numbers. 
CIFAR10 and CIFAR100 both have 50,000 training and 10,000 testing images from 10 and 100 object categories respectively.

\paragraph{Network Architectures.} We test our method on a wide range of network architectures, including multilayer perceptron (MLP), ConvNet \cite{gidaris2018dynamic}, LeNet \cite{lecun1998gradient}, AlexNet \cite{krizhevsky2012imagenet}, VGG-11 \cite{simonyan2014very} and ResNet-18 \cite{he2016deep}. 
We use the ConvNet as the default architecture in experiments unless otherwise indicated. The default ConvNet has 3 duplicate convolutional blocks followed by a linear classifier, and each block consists of 128 filters, average pooling, ReLu activation \cite{nair2010rectified} and instance normalization \cite{ulyanov2016instance}. 
We refer to \cite{zhao2021dataset} for more details about the above-mentioned architectures.
The network parameters for all architectures are randomly initialized with Kaiming initialization \cite{he2015delving}. 
The labels of synthetic data are pre-defined evenly for all classes, and the synthetic images are initialized with randomly sampled real images of corresponding class. 
While our method works well when initialising synthetic images from random noise, initialising them from randomly picked real images leads to better performance. 
We evaluate the initialization strategies in \Cref{sec:init_render}. 

\paragraph{Hyper-parameters and Augmentation.} For simplicity and generality, we use one set of hyper-parameters and augmentation strategy for all datasets.
We set $K = 1000$, $\varsigma_{\ms}=1$, $\eta_{\btheta}=0.01$, $\eta_{\ms}=0.1$, $T = 1/10/50$ and $\varsigma_{\btheta} = 1/50/10$ for 1/10/50 image(s)/class learning respectively as in \cite{zhao2021dataset}. 
The minibatch sizes for both real and synthetic data are 256.
When the synthetic set has fewer images than 256, we use all the synthetic images of a class $c$ in each minibatch.
For data augmentation, we randomly pick one of several augmentations to implement each time. More details can be found in \cref{sec:abstudy_aug_strategy}.

\paragraph{Experimental Setting.} 
We evaluate our method at three settings, 1/10/50 image(s)/class learning. 
Each experiment involves two phases. 
First, we learn to synthesize a small synthetic set (\eg 10 images/class) from a given large real training set.
Then we use the learned synthetic set to train randomly initialized neural networks and evaluate their performance on the real testing set. 
For each experiment, we learn 5 sets of synthetic images and use each set to train 20 randomly initialized networks, report mean accuracy and its standard deviation over the 100 evaluations.

\subsection{Comparison to State-of-the-Art}
\paragraph{Competitors.}
We compare our method to several state-of-the-art coreset selection and training set synthesis methods. 
The coreset selection competitors are \emph{random}, \emph{herding} \cite{chen2010super, rebuffi2017icarl, castro2018end, belouadah2020scail} and \emph{forgetting} \cite{toneva2019empirical}. 
\emph{Random} is a simple baseline that randomly select samples as the coreset. 
\emph{Herding} is a distance based algorithm that selects samples whose center is close to the distribution center, \ie each class center. 
\emph{Forgetting} is a statistics based metric that selects samples with the maximum misclassification frequencies during training. 
Training set synthesis competitors are Dataset Distillation (DD)~\cite{wang2018dataset}, Label Distillation (LD)~\cite{bohdal2020flexible}, Dataset Condensation (DC)~\cite{zhao2021dataset} which we built our method on. 
We also provide baseline performances for an approximate upperbound that are obtained by training the models on the whole real training set.
Note that we report the results of coreset selection methods and upper-bound performances presented in DC~\cite{zhao2021dataset}, as we use the same setting and hyper-parameters, and present the original results for the rest methods.

\begin{table*}
\renewcommand\arraystretch{0.9}
\centering
\scriptsize
\setlength{\tabcolsep}{3pt}
\begin{tabular}{ccc|ccc|cccc|c}
\toprule
\multirow{3}{*}{}           & \multirow{2}{*}{Img/Cls} & \multirow{2}{*}{Ratio \%} & \multicolumn{3}{c|}{Coreset Selection}   & \multicolumn{4}{c|}{Training Set Synthesis} & \multirow{2}{*}{Whole Dataset} \\ 
                            & &                & Random        & Herding       & Forgetting     & DD$^\dagger$  & LD$^\dagger$ & DC             & \emph{DSA} &       \\ \midrule
\multirow{3}{*}{MNIST}          & 1   & 0.017  & 64.9$\pm$3.5  & 89.2$\pm$1.6  & 35.5$\pm$5.6   &               & 60.9$\pm$3.2  & \bf{91.7$\pm$0.5}  & 88.7$\pm$0.6  & \multirow{3}{*}{99.6$\pm$0.0} \\
                                & 10  & 0.17   & 95.1$\pm$0.9  & 93.7$\pm$0.3  & 68.1$\pm$3.3   & 79.5$\pm$8.1  & 87.3$\pm$0.7  & 97.4$\pm$0.2  & \bf{97.8$\pm$0.1} & \\
                                & 50  & 0.83   & 97.9$\pm$0.2  & 94.8$\pm$0.2  & 88.2$\pm$1.2   & -             & 93.3$\pm$0.3  & 98.8$\pm$0.2  & \bf{99.2$\pm$0.1} &  \\ \midrule

\multirow{3}{*}{FashionMNIST}   & 1   & 0.017  & 51.4$\pm$3.8  & 67.0$\pm$1.9  & 42.0$\pm$5.5   & -             & -             & \bf{70.5$\pm$0.6}  & \bf{70.6$\pm$0.6}  & \multirow{3}{*}{93.5$\pm$0.1} \\
                                & 10  & 0.17   & 73.8$\pm$0.7  & 71.1$\pm$0.7  & 53.9$\pm$2.0   & -             & -             & 82.3$\pm$0.4  & \bf{84.6$\pm$0.3}  &           \\
                                & 50  & 0.83   & 82.5$\pm$0.7  & 71.9$\pm$0.8  & 55.0$\pm$1.1   & -             & -             & 83.6$\pm$0.4  & \bf{88.7$\pm$0.2}  &  \\ \midrule

\multirow{3}{*}{SVHN}           & 1   & 0.014  & 14.6$\pm$1.6  & 20.9$\pm$1.3  & 12.1$\pm$1.7   & -             & -             & \bf{31.2$\pm$1.4}  & 27.5$\pm$1.4  & \multirow{3}{*}{95.4$\pm$0.1} \\ 
                                & 10  & 0.14   & 35.1$\pm$4.1  & 50.5$\pm$3.3  & 16.8$\pm$1.2   & -             & -             & 76.1$\pm$0.6  & \bf{79.2$\pm$0.5}  &         \\
                                & 50  & 0.7    & 70.9$\pm$0.9  & 72.6$\pm$0.8  & 27.2$\pm$1.5   & -             & -             & 82.3$\pm$0.3  & \bf{84.4$\pm$0.4}     &  \\ \midrule

\multirow{3}{*}{CIFAR10}        & 1   & 0.02   & 14.4$\pm$2.0  & 21.5$\pm$1.2  & 13.5$\pm$1.2   & -             & 25.7$\pm$0.7  & \bf{28.3$\pm$0.5}  & \bf{28.8$\pm$0.7}  & \multirow{3}{*}{84.8$\pm$0.1} \\
                                & 10  & 0.2    & 26.0$\pm$1.2  & 31.6$\pm$0.7  & 23.3$\pm$1.0   & 36.8$\pm$1.2  & 38.3$\pm$0.4  & 44.9$\pm$0.5  & \bf{52.1$\pm$0.5}  &           \\  
                                & 50  & 1      & 43.4$\pm$1.0  & 40.4$\pm$0.6  & 23.3$\pm$1.1   & -             & 42.5$\pm$0.4  & 53.9$\pm$0.5  & \bf{60.6$\pm$0.5}     &  \\ \midrule
                                
\end{tabular}
\caption{\footnotesize{The performance comparison to coreset selection and training set synthesis methods. This table shows the testing accuracies (\%) of models trained from scratch on the small coreset or synthetic set. Img/Cls: image(s) per class, Ratio~(\%): the ratio of condensed images to whole training set. DD$^\dagger$ and LD$^\dagger$ use LeNet for MNIST and AlexNet for CIFAR10, while the rest use ConvNet for training and testing.}}
\label{tab:sota_small}
\vspace{-15pt}
\end{table*}

\paragraph{Results for 10 Category Datasets.} \Cref{tab:sota_small} presents the results of different methods on MNIST, FashionMNIST, SVHN and CIFAR10, which all have 10 classes. 
We first see that \emph{Herding} performs best among the coreset methods for a limited number of images \eg only 1 or 10 image(s)/class and \emph{random} selection performs best for 50 images/class.
Overall the training set synthesis methods outperform the coreset methods which shows a clear advantage of synthesizing images for the downstream tasks, especially for 1 or 10 image(s)/class.
Our method achieves the best performance in most settings and in particular obtains significant gains when learning 10 and 50 images/class, improves over the state-of-the-art DC by 7.2\% and 6.7\% in CIFAR10 dataset for 10 and 50 images per class.
Remarkably in MNIST with less than 1\% data (50 images/class), it achieves 99.2\% which on par with the upperbound 99.6\%.
We also observe that our method obtains comparable or worse performance than DC in case of 1 image/class.
We argue that our method acts as a regularizer on DC, as the synthetic images are forced to match the gradients from real training images under different transformations.
Thus we expect that our method works better when the solution space (synthetic set) is larger.
Finally, the performance gap between the training set synthesis methods and upperbound gets larger when the task is more challenging.
For instance, in the most challenging dataset CIFAR10, the gap between ours and the upperbound is 24.2\%, while it is 0.4\% in MNIST in the 50 images/class setting.


Note that we are aware of two recent work, Generative Teaching Networks (GTN)~\cite{such2020generative} and  Kernel Inducing Point (KIP)~\cite{nguyen2021dataset}.
GTN provides only their performance curve on MNIST for 4,096 synthetic images ($\approx$ 400 images/class) but no numerical results, which is slightly worse than our performance with 50 images/class. 
KIP achieves 95.7$\pm$0.1\% and 46.9$\pm$0.2\% testing accuracies on MNIST and CIFAR10 when learning 50 images/class with kernels and testing with one-layer fully connected network, while our results with ConvNet are 99.2$\pm$0.1\% and 60.6$\pm$0.5\% respectively.
Though our results are significantly better than theirs, two methods are not directly comparable, 
as KIP and our DSA use different training and testing architectures.

\begin{figure}[]
    \centering
    \includegraphics[width=1.0\linewidth]{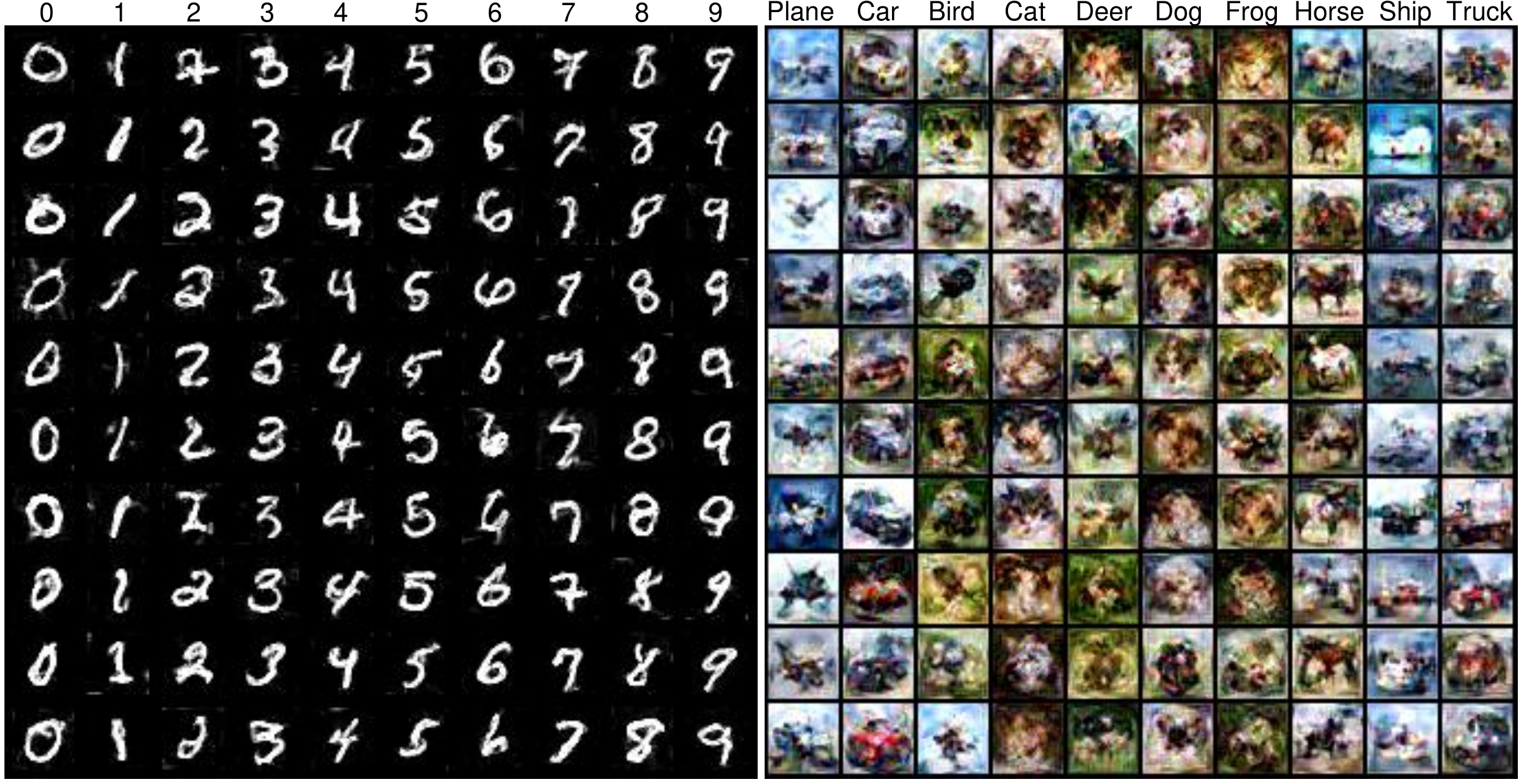}
    \caption{\footnotesize{Visualization of the generated 10 images/class synthetic sets of MINIST and CIFAR10. The synthetic images are easy to recognize for human beings.}}
    \label{fig:vis}
    \vspace{-15pt}
\end{figure}

We visualize the generated 10 images/class synthetic sets of MNIST and CIFAR10 in \Cref{fig:vis}. 
Overall the synthetic images capture diverse appearances in the categories, various writing styles in MNIST and a variety of viewpoints and background in CIFAR10.
Although it is not our goal, our images are easily recognizable and more similar to real ones than the ones that are reported in \cite{wang2018dataset, such2020generative, nguyen2021dataset}.


\paragraph{CIFAR100 Results.} We also evaluate our method in the more challenging CIFAR100 dataset in which few works report their results.
Note that compared to CIFAR10, CIFAR100 is more challenging, as recognizing 10 times more categories requires to learn more powerful features and also there are $\frac{1}{10}$ fewer images per class in CIFAR100.
We present our results in \Cref{tab:sota_large} and compare to the competitive coreset methods (\emph{random}, \emph{herding}) and train set synthesis methods (\emph{LD}, \emph{DC}).
Our method obtains 13.9\% and 32.3\% testing accuracies for 1 and 10 images/class, which improves over DC by 1.1\% and 7.1\% respectively.
Compared to the 10 category datasets, the relative  performance gap between the upperbound (56.2$\pm$0.3\%) and the best performing method (DSA) is significantly bigger in this dataset.


\begin{table}[]
\renewcommand\arraystretch{0.9}
\centering
\scriptsize
\setlength{\tabcolsep}{2pt}
\begin{tabular}{c|cc|ccc|c}
\toprule
Img/Cls & Random        & Herding   & LD$^\dagger$ & DC            & \emph{DSA}     & Whole Dataset\\ \midrule
1       &  4.2$\pm$0.3  &  8.4$\pm$0.3    & 11.5$\pm$0.4  & 12.8$\pm$0.3  & \bf{13.9$\pm$0.3}        & \multirow{2}{*}{56.2$\pm$0.3}\\
10      & 14.6$\pm$0.5  & 17.3$\pm$0.3    & -             & 25.2$\pm$0.3  & \bf{32.3$\pm$0.3}  & \\ \bottomrule     
\end{tabular}
\caption{\footnotesize{The performance (\%) comparison on CIFAR100 dataset. LD$^\dagger$ use AlexNet for CIFAR100, while the rest use ConvNet.}}
\label{tab:sota_large}
\vspace{-15pt}
\end{table}


\begin{table}[]
\centering
\scriptsize
\setlength{\tabcolsep}{3pt}
\begin{tabular}{ccccccc}
	\toprule
\texttt{C}\textbackslash \texttt{T} & MLP           & ConvNet       & LeNet             & AlexNet       & VGG               & ResNet\\ 	\midrule
	MLP                             & 76.5$\pm$1.2  & 73.1$\pm$3.6  & 80.4$\pm$2.9  & 78.2$\pm$6.4  & 58.7$\pm$6.5  & 78.7$\pm$3.9  \\
	ConvNet                         & 75.6$\pm$1.1  & 88.8$\pm$0.8  & 84.8$\pm$1.5  & 84.7$\pm$1.5  & 83.5$\pm$1.7  & 89.3$\pm$0.8   \\ 
	LeNet                           & 76.5$\pm$0.9  & 86.6$\pm$1.5  & 83.9$\pm$1.6  & 83.9$\pm$1.2  & 81.1$\pm$2.3  & 88.2$\pm$0.9   \\ 
	AlexNet                         & 76.1$\pm$0.8  & 87.6$\pm$0.8  & 84.2$\pm$1.6  & 84.6$\pm$1.7  & 82.0$\pm$2.1  & 88.8$\pm$0.8   \\ 
	VGG                             & 75.8$\pm$1.0  & 88.9$\pm$0.7  & 84.5$\pm$1.6  & 85.0$\pm$1.4  & 83.2$\pm$1.9  & 88.9$\pm$1.0   \\ 
	ResNet                          & 75.8$\pm$1.0  & 88.6$\pm$0.8  & 84.8$\pm$1.7  & 84.8$\pm$1.2  & 82.4$\pm$1.9  & 89.5$\pm$1.0   \\ \bottomrule
\end{tabular}
\caption{\footnotesize{Cross-architecture performance (\%). We learn to condense the training set on one architecture (C), and test it on another architecture (T). We learn 1 image/class condensed set on MNIST.}}
\label{tab:crossarc}
\vspace{-15pt}
\end{table}

\subsection{Cross-Architecture Generalization}
Here we study the cross-architecture performance of our model and report the results in \Cref{tab:crossarc} in MNIST for 1 image/class.
To this end, we use different neural network architectures to learn the synthetic images and further use them to train classifiers.
The rows indicate the architecture which is used to learn the synthetic images and columns show the architectures that we train classifiers.
The results show that synthetic images learned by the convolutional architectures (ConvNet, LeNet, AlexNet, VGG and ResNet) perform best and generalizes to the other convolutional ones, while the MLP network produces less informative synthetic images overall.
Finally the most competitive architecture, ResNet provides the best results when trained as a classifier on the synthetic images.


\subsection{Ablation Study}
\label{sec:ablation_study}
\paragraph{Effectiveness of DSA.}
Here we study the effect of design choices in the proposed DSA in terms of test performance on CIFAR10 for 10 images/class and report it in \Cref{tab:abstudy_effectiveness}.
One can apply image transformations to the real and synthetic set while learning the synthetic images, also to the synthetic images while training a classifier in the second stage.
In addition, the same image transformation can be applied to all images in a real and synthetic minibatch pair (denoted as $\ma_{\omega}$) or an independently sampled image transformation can be applied to each image (denoted as $\ma$).
Note that the former corresponds to our proposed Siamese augmentation and we test different augmentation schemes for cropping, flipping, scaling and rotation.

The results verify that the proposed Siamese augmentation always achieves the best performance when used with individual augmentation. The largest improvement is obtained by applying our Siamese augmentation with cropping. 
Specifically, using Siamese augmentation with cropping achieves $3.6\%$ improvement compared to no data augmentation (A). 
Note that (A) corresponds to DC \cite{zhao2021dataset} with initialization from real images.
While smaller improvement of 1.4\% can be obtained by applying cropping only to synthetic data in test phase (B), DSA provides a further 2.2\% over this.
Applying cropping only to the real or synthetic images (C and D) degrades the performance and obtains worse performance than no data augmentation (A).
Similarly, applying independent transformations to the real and synthetic images when learning synthetic images, \ie (F), leads to worse performance than (A).
Finally, the Siamese augmentation method performs worse than (A) when no data augmentation is used to train the classifier (E).
This shows that it is important to apply data augmentation consistently in both stages. 
The effects on other augmentation strategies may be slightly different but similar to those on cropping augmentation.



\begin{table}[]
\centering
\scriptsize
\setlength{\tabcolsep}{3pt}
\begin{tabular}{cccccccc}
\toprule
     & \multicolumn{2}{c}{Condense}    & Test      & \multicolumn{4}{c}{Test Performance (\%)}                                 \\
     & Real           & Synthetic      & Synthetic & Crop             & Flip             & Scale            & Rotation         \\ \midrule
Ours & $\ma_{\omega}$ & $\ma_{\omega}$ & $\ma$     & 49.1$\pm$0.6 & 47.9$\pm$0.7 & 46.9$\pm$0.5 & 46.8$\pm$0.6 \\ \midrule
(A)  & -              & -              & -         & 45.5$\pm$0.6     & 45.5$\pm$0.6     & 45.5$\pm$0.6     & 45.5$\pm$0.6     \\
(B)  & -              & -              & $\ma$     & 46.9$\pm$0.6     & 46.1$\pm$0.6     & 45.7$\pm$0.5     & 45.0$\pm$0.5     \\
(C)  & $\ma$          & -              & $\ma$     & 42.8$\pm$0.7     & 46.2$\pm$0.6     & 44.5$\pm$0.6     & 44.5$\pm$0.6     \\
(D)  & -              & $\ma$          & $\ma$     & 44.6$\pm$0.7     & 46.8$\pm$0.6     & 45.4$\pm$0.6     & 45.9$\pm$0.7     \\
(E)  & $\ma_{\omega}$ & $\ma_{\omega}$ & -         & 43.4$\pm$0.5     & 46.4$\pm$0.6     & 45.7$\pm$0.6     & 46.3$\pm$0.5     \\
(F)  & $\ma$          & $\ma$          & $\ma$     & 44.5$\pm$0.5     & 46.9$\pm$0.6     & 45.7$\pm$0.5     & 45.8$\pm$0.5     \\ \bottomrule
\end{tabular}
\caption{\footnotesize{Ablation study on augmentation schemes in CIFAR10 for 10 synthetic images/class. $\ma_{\omega}$ denotes Siamese augmentation when applied to both real and synthetic data, while $\ma$ denotes augmentation that is not shared across real and synthetic minibatches.}}
\label{tab:abstudy_effectiveness}
\vspace{-15pt}
\end{table}

\paragraph{Augmentation Strategy.}
\label{sec:abstudy_aug_strategy}
Our method can be used with the common image transformations. 
Here we investigate the performance of our method with several popular transformations including color jittering, cropping, cutout, flipping, scaling, rotation on MNIST, FashionMNIST, SVHN and CIFAR10 for 10 images/class setting.
We also show a simple combination strategy that is used as the default augmentation in experiments by randomly sampling one of these six transformations at each time. 
The exception is that flipping is not included in the combination for the two number datasets - MNIST and SVHN, as it can change the semantics of a number.
Note that our goal is not to exhaustively find the best augmentation strategy but to show that our augmentation scheme can be effectively used for dataset condensation and we leave a more elaborate augmentation strategy for future work.

\Cref{tab:abstudy_strategy} depicts the results for no transformation, individual transformations and as well as the combined strategy.
We find that applying all the augmentations improve the performance on CIFAR10 compared to the baseline (None). 
Cropping is the most effective single transformation that can increase the testing accuracy from 45.5\% to 49.1\%. 
The combination of these augmentations further improves the performance to 52.1\%. 
Interestingly cropping and cutout transformations degrade the performance of SVHN, as SVHN images are noisy and some include multiple digits and these transformations may pick the wrong patch of images.
Nevertheless, we still observe that the combined strategy obtains the best performance in all datasets.

\begin{table*}[]
\centering
\scriptsize
\setlength{\tabcolsep}{4pt}
\begin{tabular}{cccccccccc}
\toprule
       & Img/Cls            & None          & Color         & Crop          & Cutout        & Flip          & Scale         & Rotate        & Combination  \\ \midrule
MNIST & 10                  & 96.4$\pm$0.1  & 96.5$\pm$0.1  & 97.2$\pm$0.1 & 96.5$\pm$0.1   & -             & 97.2$\pm$0.1  & 97.2$\pm$0.1  & 97.8$\pm$0.1  \\ 
FashionMNIST & 10           & 82.5$\pm$0.3  & 82.9$\pm$0.3  & 83.3$\pm$0.3 & 84.0$\pm$0.3   & 83.1$\pm$0.2  & 84.0$\pm$0.4  & 83.1$\pm$0.3  & 84.6$\pm$0.3   \\
SVHN & 10                   & 76.7$\pm$0.6  & 77.4$\pm$0.5  & 75.9$\pm$0.8  & 73.1$\pm$0.6  & -             & 78.0$\pm$0.5  & 77.4$\pm$0.4  & 79.2$\pm$0.5   \\ 
CIFAR10 & 10                & 45.5$\pm$0.6  & 47.6$\pm$0.5  & 49.1$\pm$0.6  & 48.1$\pm$0.5  & 47.9$\pm$0.7  & 46.9$\pm$0.5  & 46.8$\pm$0.6  & 52.1$\pm$0.5   \\
\bottomrule
\end{tabular}
\caption{\footnotesize{Performance with different augmentation strategies. Flipping is not suitable for number datasets - MNIST and SVHN. Combination means randomly sampling one from the six/five transformations to implement each time.}}
\label{tab:abstudy_strategy}
\vspace{-12pt}
\end{table*}


\subsection{Continual Learning}
Here we apply our method to a continual learning task \cite{rebuffi2017icarl, castro2018end, aljundi2019gradient} where the tasks are incrementally learned on three digit recognition datasets, SVHN \cite{netzer2011reading}, MNIST \cite{lecun1998gradient} and USPS \cite{hull1994database} as in \cite{zhao2021dataset} and the goal is to preserve the performance in the seen tasks while learning new ones.
We build our model on the popular continual learning baseline -- EEIL~\cite{castro2018end} which leverages memory rehearsal and knowledge distillation \cite{hinton2015distilling} to mitigate catastrophic forgetting of old tasks. We replace the sample selection strategy, \ie herding, with our dataset condensation method for memory construction and keep the rest the same. The memory budget is 10 images/class for all seen classes. We refer to \cite{zhao2021dataset} for more details.

\begin{figure}[]
    \centering
    \includegraphics[width=0.9\linewidth]{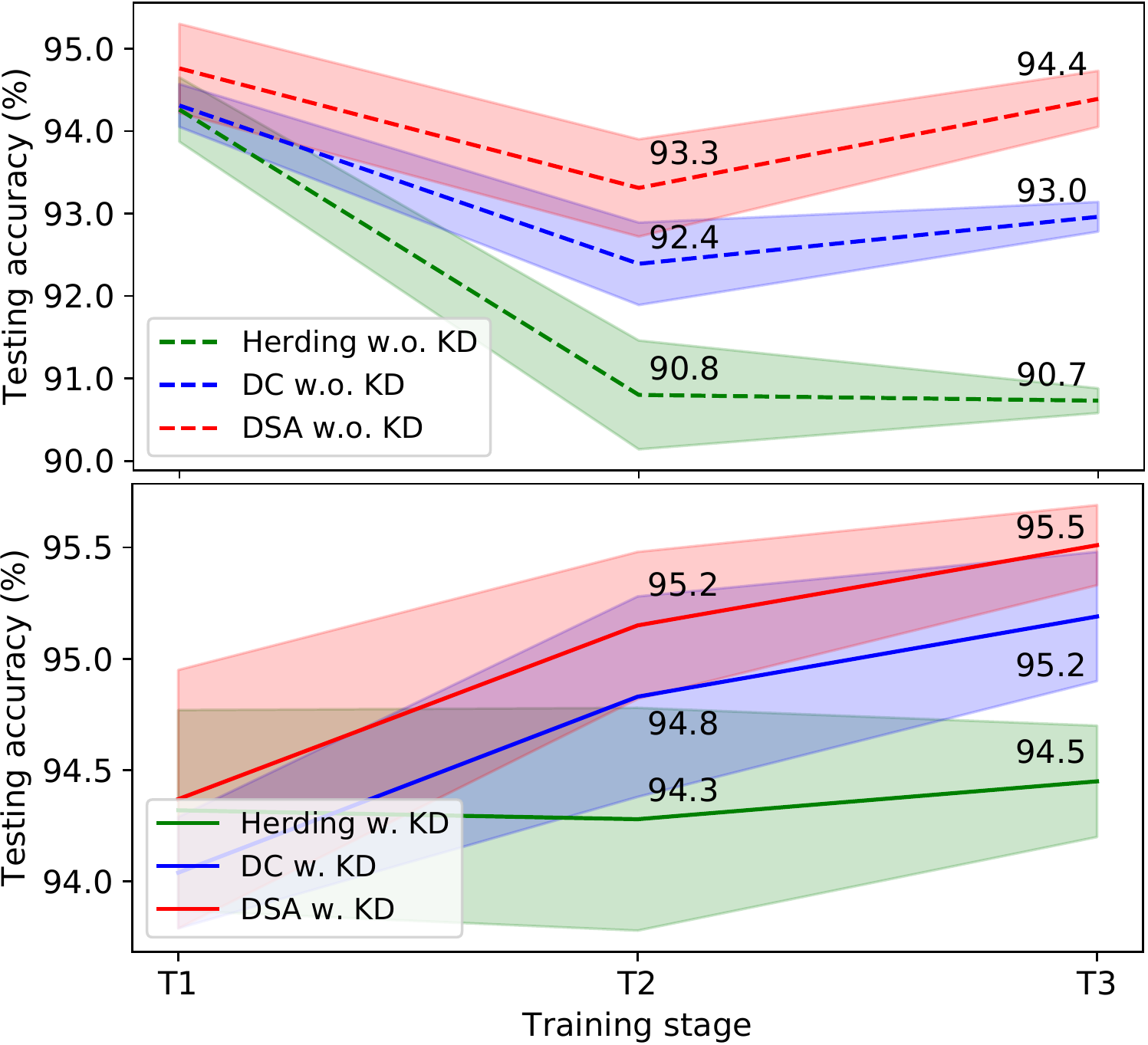}
    \caption{\footnotesize{Continual learning performance. We compare to the original EEIL \cite{castro2018end} denoted as Herding and DC \cite{zhao2021dataset} under two settings: with and without knowledge distillation. T1, T2, T3 are three learning stages.}}
    \label{fig:cl}
    \vspace{-15pt}
\end{figure}

\Cref{fig:cl} depicts the results of EEIL with three memory construction strategies - herding \cite{castro2018end}, DC \cite{zhao2021dataset} and our DSA under two settings - with and without knowledge distillation. The results show that DSA always outperforms the other two memory construction methods. Especially, DSA achieves 94.4\% testing accuracy after learning all three tasks without knowledge distillation, which surpasses DC and herding by 1.4\% and 3.7\% respectively. It indicates that the synthetic images learned by our method are more informative for training models than those produced by competitors.

\begin{figure}[]
    \centering
    \includegraphics[width=0.95\linewidth]{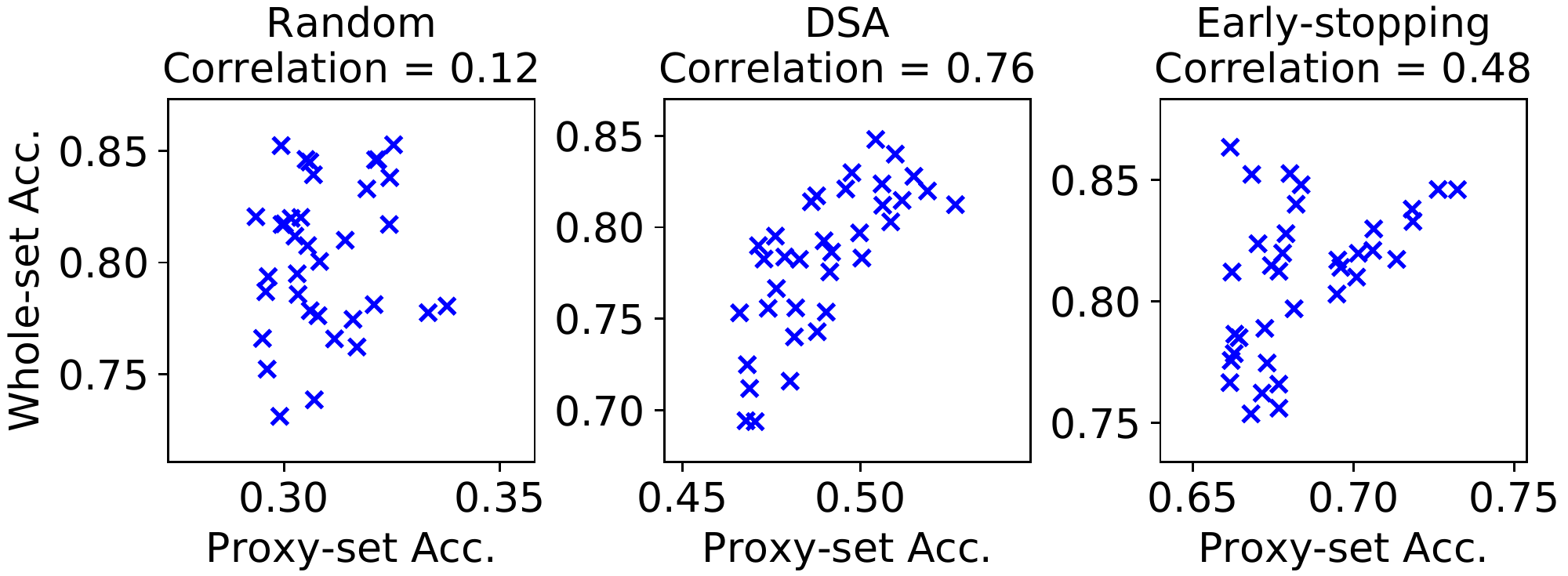}
    \caption{\footnotesize{The distribution of correlation between proxy-set performance and whole-dataset performance on top 5\% architectures.}}
    \label{fig:nas}
    \vspace{-15pt}
\end{figure}

\subsection{Neural Architecture Search}
Our method has substantial practical benefits when one needs to train many neural networks in the same dataset.
One such application is neural architecture search (NAS)~\cite{zoph2018learning} which aims to search the best network architecture for a given dataset.
Here our goal is to verify that a small set of synthetic images learned by our method can be used as a proxy set to efficiently train many networks and the test performances of the neural architectures are correlated to the ones trained on the original training set.

To this end, we design a set of candidate neural network architectures based on the modular ConvNet by varying its depth, width, pooling, activation and normalization layers which produces 720 candidates in total.
We refer to \cite{zhao2021dataset} for more details.  
We train these models on the whole CIFAR10 training set for obtaining the ground-truth performance and also three small proxy-sets that are obtained with \emph{random}, \emph{DSA} and \emph{early-stopping} \cite{li2020random}. 
We randomly select 10 images/class in \emph{random} and learn 10 images/class condensed set with the default ConvNet in \emph{DSA} as the proxy-sets. 
We train models 300 epochs in \emph{random} and \emph{DSA}. 
In \emph{early-stopping} we train models for same amount of iterations to DSA (300 iterations with batch size 100) on the whole original training set.
Both \emph{random} and \emph{DSA} use 100 images (0.2\% of whole dataset) in total, while \emph{early-stopping} uses $3\times 10^4$ images (60\% of whole dataset). 
For the whole set baseline, we train models for 100 epochs, which is sufficiently long to converge.
Finally we pick the best performing architectures that are trained on each proxy set, train them on the original training set from scratch and report their test set performance.

We report the results in \Cref{tab:nas} in the performance of selected best architecture, correlation between performances of proxy-set and whole-dataset training, training time cost and storage cost. 
The correlation, \ie Spearman’s rank correlation coefficient, is calculated on the top 5\% candidate architectures of each proxy-set, which is also illustrated in Figure \ref{fig:nas}. 
The proxy-set produced by our method achieves the strongest correlation - 0.76, while the time cost of implementing NAS on our proxy-set is only 1.2\% of implementing NAS on the whole dataset.
This promising performance indicates that our DSA can speed up NAS by training models on small proxy-set. 
Although the model chosen by \emph{early-stopping} achieves better performance than ours, \emph{early-stopping} requires two orders of magnitude more training images than ours. 
In addition, the correlation (0.48) between \emph{early-stopping} performance and whole-dataset training performance is significantly lower than ours (0.76).

\begin{table}[]
\renewcommand\arraystretch{0.9}
\centering
\scriptsize
\setlength{\tabcolsep}{3pt}
\begin{tabular}{cccc|c}
\toprule
                    & Random            & DSA            & Early-stopping    & Whole Dataset \\ \midrule
Performance (\%)    & 78.2              & 81.3              & \textbf{84.3}              & 85.9  \\
Correlation         & 0.12              & \textbf{0.76}     & 0.48              & 1.00 \\
Time cost (min)     & \textbf{32.5}     & \textbf{44.5}     & \textbf{32.6}     & 3580.2 \\ 
Storage (imgs)      & $\mathbf{10^2}$   & $\mathbf{10^2}$   & $3\times 10^4$            & $5\times 10^4$   \\ \bottomrule
\end{tabular}
\caption{\footnotesize{Neural architecture search on proxy-sets and whole dataset. The search space is 720 ConvNets. We do experiments on CIFAR10 with 10 images/class randomly selected coreset and synthetic set learned by \emph{DSA}. \emph{Early-stopping} means training models on whole dataset but with the same iterations as \emph{random} and \emph{DSA}.}}
\label{tab:nas}
\vspace{-15pt}
\end{table}


\begin{figure}[]
    \centering
    \includegraphics[width=0.79\linewidth]{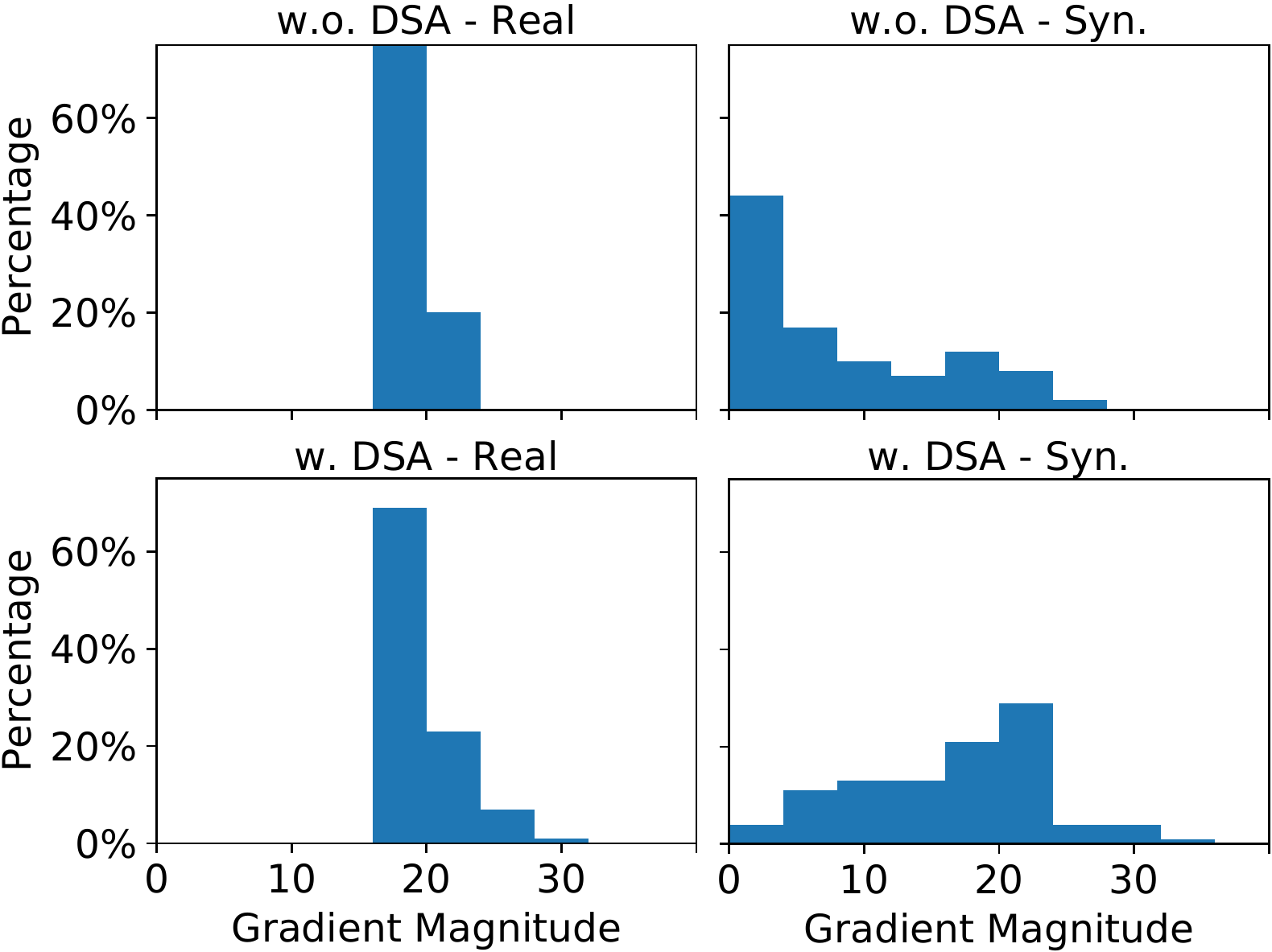}
    \caption{\footnotesize{Gradient magnitude distribution w.r.t. real/synthetic data.}}
    \label{fig:grad_distribution}
    \vspace{-15pt}
\end{figure}

\section{Discussion}
\label{sec:discussion}
\subsection{Why Does DSA Work?}
In this section, we attempt to shed light onto why DSA leads to better synthetic data.
We hypothesize that the Siamese augmentation acts as a strong regularizer on the learned high-dimensional synthetic data $\ms$ and alleviates its overfitting to the real training set.
We refer to \cite{hernandez2018data} for more elaborate analysis of the relation between data augmentation and regularization.
This can be shown more explicitly by reformulating \cref{eq.optgradAug} over multiple randomly sampled augmentations: 
\begin{equation}
    \min_{\ms} \sum_{\omega \sim \Omega} D(\nabla_{\btheta}\ml(\ma(\ms,\omega), \btheta),\nabla_{\btheta}\ml(\ma(\mt,\omega), \btheta)),
    \label{eq.optgradConst}
\end{equation} which forces the synthetic set to match the gradients from the real set under multiple transformations $\omega$ when sampled from the distribution $\Omega$ and renders the optimization harder and less prone to overfitting.

We also quantitatively analyze this by reporting the gradient magnitude distribution $\lvert\lvert \nabla_{\btheta}\ml(\mt)\rvert\rvert$ and $\lvert\lvert \nabla_{\btheta}\ml(\ms)\rvert\rvert$ for real and synthetic sets respectively in \Cref{fig:grad_distribution} when learning 10 images/class synthetic set on CIFAR10 with and without DSA.
The gradients are obtained at the training iteration $k = 1000$ (see Alg.~\ref{algo}).
We see that the gradient magnitudes from the synthetic data quickly vanishes and thus leads to a very small updates in absence of DSA, while the synthetic images can still be learned with DSA.
Note that as backpropagation involves successive products of gradients, the updates for $\ms$ naturally vanishes when multiplied with small $\lvert\lvert \nabla_{\btheta}\ml(\ms)\rvert\rvert$.

\subsection{Initializing Synthetic Images}
\label{sec:init_render}
\begin{figure}[]
    \centering
    \includegraphics[width=1\linewidth]{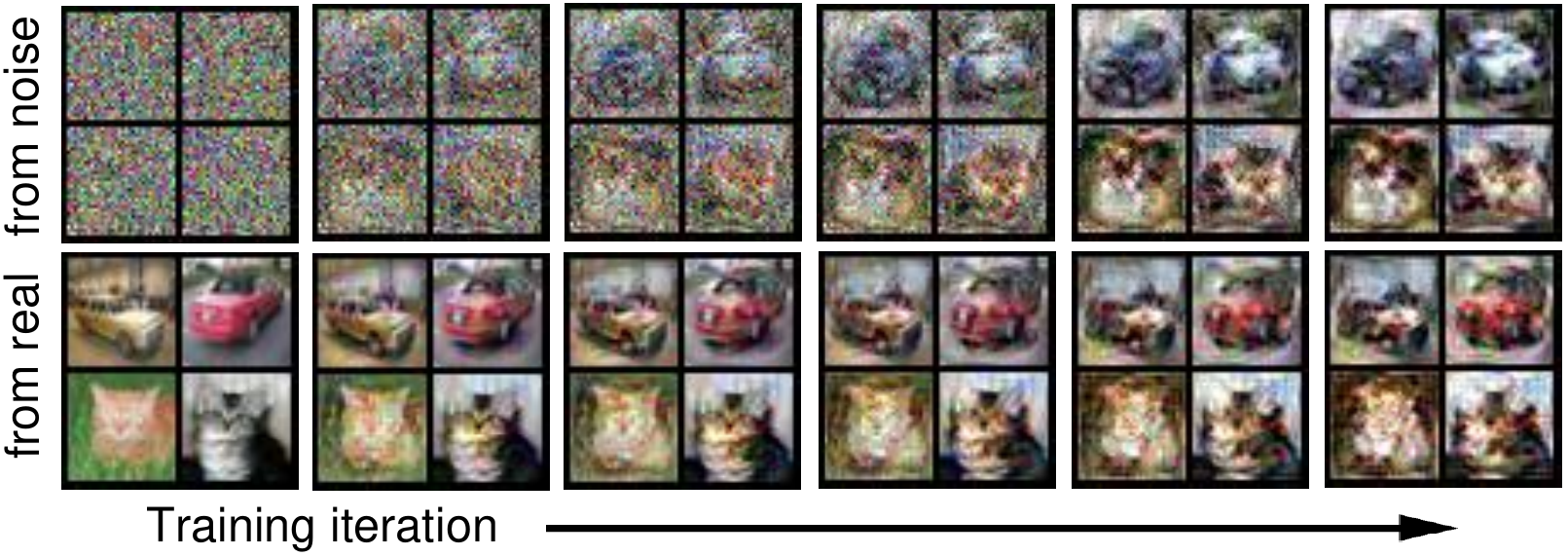}
    \caption{\footnotesize{The learning/rendering process of two classes in CIFAR10 initialized from random noise and real images respectively.}}
    \label{fig:rendering}
    \vspace{-15pt}
\end{figure}

In our experiments, we initialize each synthetic image with a randomly sampled real training image (after standard image normalization) from the corresponding category.
After the initialization, we update them by using the optimization in \cref{eq.optgradAug}.
Once they are trained, they are used to train neural networks without any post-processing.
In \Cref{fig:rendering}, we illustrate the evolution of synthetic data initialized from random noise and real images from car and cat categories through our training in CIFAR10. 
While we see significant changes over the initialization in both cases, the ones initialized with real images preserve some of their contents such as object pose and color.

\section{Conclusion}
\label{sec:conclusion}
In this paper, we propose a principled dataset condensation method -- Differentiable Siamese Augmentation -- to enable learning synthetic training set that can be effectively used with data augmentation when training deep neural networks. 
Experiments and ablation study show that the learned synthetic training set can be used to train neural networks with data augmentation and achieve significantly better performance (about 7\% improvement on CIFAR10/100) than state-of-the-art methods. 
We also show promising results when applying the proposed method to continual learning and neural architecture search.

\paragraph{Acknowledgment.} This work is funded by China Scholarship Council 201806010331 and the EPSRC programme grant Visual AI EP/T028572/1.

\bibliography{refs}
\bibliographystyle{icml2021}

\end{document}